\documentclass{article}
\usepackage{PRIMEarxiv}

\usepackage[utf8]{inputenc} 
\usepackage[T1]{fontenc}    
\usepackage{hyperref}       
\usepackage{url}            
\usepackage{booktabs}       
\usepackage{amsfonts}       
\usepackage{nicefrac}       
\usepackage{microtype}      
\usepackage{lipsum}
\usepackage{fancyhdr}       
\usepackage{graphicx}       
\graphicspath{{media/}}     

\usepackage{xcolor, soul, tikz, svg, subcaption, enumitem, adjustbox, float}
\usepackage{array, caption, makecell, multirow, tabularx, tcolorbox, verbatim}
\usepackage{amsthm, amsmath,amssymb,amsfonts, mathrsfs, textcomp, manyfoot}
\usepackage{algorithm, algpseudocode, listings, siunitx}

\definecolor{lime}{HTML}{A6CE39}
\DeclareRobustCommand{\orcidicon}{
	\begin{tikzpicture}
	\draw[lime, fill=lime] (0,0) 
	circle [radius=0.16] 
	node[white] {{\fontfamily{qag}\selectfont \tiny ID}};
	\draw[white, fill=white] (-0.0625,0.095) 
	circle [radius=0.007];
	\end{tikzpicture}
	\hspace{-2mm}
}
\foreach \x in {A, ..., Z}{\expandafter\xdef\csname orcid\x\endcsname{\noexpand\href{https://orcid.org/\csname orcidauthor\x\endcsname}
			{\noexpand\orcidicon}}
}

\title{ExBigBang: A Dynamic Approach for Explainable Persona Classification through Contextualized  Hybrid Transformer Analysis}

\author{
    Saleh Afzoon$^{1}$ \orcidA{}, 
    Amin Beheshti$^{1}$ \orcidB{}, 
    Nabi Rezvani$^{1}$ \orcidC{},
    Farshad Khunjush$^{2}$ \orcidD{},
    Usman Naseem$^{1}$ \orcidE{},\\[0.5em]
    \textbf{John McMahon$^{3}$ \orcidF{}, 
    Zahra Fathollahi$^{1}$ \orcidG{}, 
    Mahdieh Labani$^{1}$ \orcidH{}, 
    Wathiq Mansoor$^{4}$ \orcidI{},
    Xuyun Zhang$^{1}$ \orcidJ{}}\\
    $^{1}$School of Computing, Macquarie University, Sydney, Australia \\
    $^{2}$Department of Computer Engineering and Information Technology, Shiraz University of Technology, Shiraz, Iran \\
    $^{3}$John Walsh Centre for Rehabilitation Research, Kolling Institute, University of Sydney, Sydney, Australia \\
    $^{4}$College of Engineering and Information Technology, University of Dubai, Dubai, United Arab Emirates \\
    \small \texttt{\{saleh.afzoon, nabiallah.rezvani, mahdieh.labani\}@hdr.mq.edu.au}, \small \texttt{khunjush@shirazu.ac.ir},\\ \small \texttt{ \{amin.beheshti, usman.naseem, zahra.fathollahi\_chaleshtori, xuyun.zhang \}@mq.edu.au},\\ \small \texttt{wmansoor@ud.ac.ae}
}

\begin{document}
\maketitle

\begin{abstract}
In user-centric design, persona development plays a vital role in understanding user behaviour, capturing needs, segmenting audiences, and guiding design decisions. However, the growing complexity of user interactions calls for a more contextualized approach to ensure designs align with real user needs. While earlier studies have advanced persona classification by modelling user behaviour, capturing contextual information, especially by integrating textual and tabular data, remains a key challenge. These models also often lack explainability, leaving their predictions difficult to interpret or justify. To address these limitations, we present ExBigBang (Explainable BigBang), a hybrid text-tabular approach that uses transformer-based architectures to model rich contextual features for persona classification. ExBigBang incorporates metadata, domain knowledge, and user profiling to embed deeper context into predictions. Through a cyclical process of user profiling and classification, our approach dynamically updates to reflect evolving user behaviours. Experiments on a benchmark persona classification dataset demonstrate the robustness of our model. An ablation study confirms the benefits of combining text and tabular data, while Explainable AI techniques shed light on the rationale behind the model's predictions.
\end{abstract}


\section{Introduction}\label{sec1}

Understanding user preferences is increasingly vital in today’s competitive digital landscape, where data-driven personalization benefits both businesses and end-users alike \cite{beheshti2020towards}.
Effectively capturing the complexity of user interactions provides a strategic advantage by enabling personalized experiences and improving engagement. Achieving this alignment with user needs, goals, and preferences relies on in-depth behavioural analysis.
Given the diversity of users and the multifaceted nature of their interactions, personalization must reflect the specific contexts in which users engage. A contextualized approach enhances relevance and supports the success of digital platforms.

The assumption that users with similar personas share comparable needs forms the foundation of persona analysis, revealing latent patterns and user-specific requirements \cite{konstantakis2020personalized}. Traditionally, personas were developed manually through interviews and surveys \cite{banerjee2004face, lindgren2007using, miaskiewicz2008latent}. The field then evolved with the adoption of quantitative methods \cite{goodman2018evaluating}, enabling large-scale persona construction and early work in Automatic Personality Detection (APD) \cite{tu2010combine, kim2019innocent}. Machine learning (ML) further advanced persona development through clustering \cite{hou2020method, salminen2020enriching} and classification techniques \cite{pattisapu2017medical, kaul2020persona, bharadwaj2018persona}, supporting more scalable solutions.
In parallel, user profiling emerged as a way to model behavioural patterns over time and tailor services to specific contexts \cite{eke2019survey}. Initial approaches relied on features like TF-IDF and POS tagging to infer traits \cite{park2015automatic}. Later work applied kernel density estimation for emotional profiling \cite{bernabe2015emotional}, while recent techniques use deep-contextualized semantic models and multi-layer co-occurrence networks for tasks like bot detection \cite{heidari2020deep}.
Despite these developments, constructing a comprehensive understanding of users remains challenging. Contextualizing raw data—through metadata, domain knowledge, and profile features—requires careful feature engineering to preserve meaning. Another issue is the lack of interpretability in many persona development models. Their black-box nature reduces transparency, making the integration of Explainable AI (XAI) essential for detecting bias and building trust.
A promising direction lies in unifying profiling and persona classification within a dynamic, cyclical system. Here, insights from one process continually enhance the other, enabling models to reflect behavioural changes and improve persona relevance.

While prior research has improved persona classification, most studies overlook full integration of contextualization, interpretability, and feedback between profiling and persona modelling—limiting their adaptability to real-world dynamics.
This paper addresses these limitations by introducing a contextualized persona development approach using a hybrid (text-tabular) transformer model that unifies profiling and classification. By incorporating metadata, domain knowledge, and profile-driven features, our method improves both performance and transparency. We apply XAI to explain feature influence and conduct an ablation study to assess the impact of contextual integration.
More specifically, the main contributions of this paper are:
\begin{itemize}
    \item [$\bullet$] We propose a hybrid contextualization framework for persona development using a text-tabular transformer classifier.
    \item [$\bullet$] We introduce a dynamic, cyclical integration of profiling and classification for iterative persona refinement.
    \item [$\bullet$] We apply XAI to identify and interpret the contextual features influencing persona outcomes.
\end{itemize}

The rest of the paper is structured as follows: Section~\ref{sec2} reviews related work. Section~\ref{sec3} presents our proposed method, while Section~\ref{sec4} discusses experimental results. Finally, Section~\ref{sec5} concludes the study.

\section{Related Works}\label{sec2}

In this section, we review relevant studies that contribute to the design of different components in our proposed hybrid persona development approach, focusing on techniques and challenges.

\subsection{Profiling techniques}

Data profiling serves multiple purposes, ranging from improving data quality and detecting errors to enabling a deeper understanding of the data. Such understanding is essential for developing XAI frameworks that enhance transparency and build trust, making AI systems more suitable for real-world decision-making.

\subsubsection{Statistical techniques}

In the context of user profiling, statistical methods are among the earliest approaches used to interpret user behaviour and preferences in an explainable manner.

A time-based interaction profiling approach was used to monitor software usage through system logs and generate administrator-level alerts \cite{corney2011detection}. While descriptive statistics provided useful insights, the method lacked scalability under high load and did not incorporate task context, leading to false-positive alerts.
A fixed analytical framework was applied to analyze urban behaviours using mobile phone call detail records \cite{phithakkitnukoon2010activity}. Although activity-aware maps clearly represented current and past trends, the framework lacked predictive capabilities and adaptability across different datasets.
Multivariate Kernel Density Function (MKDF) was employed to represent long-term emotional states in a two-dimensional space of polarity and intensity \cite{bernabe2015emotional}. While MKDF offers interpretability through simplicity, it struggles with capturing complex patterns, scaling to large datasets, and providing depth comparable to advanced machine learning techniques.

\subsubsection{Filtering methods}

Profiling methods have evolved to accommodate users' complex and dynamic preferences. Filtering techniques, unlike statistical ones, are better at capturing this complexity through adaptive learning, leading to more accurate recommendations. Rule-based profiling, which uses static demographic information and predefined rules, offers transparency and easy domain integration. However, self-defined rules can introduce biases and are difficult to maintain \cite{eke2019survey}.
Content-based filtering provides personalized recommendations by analyzing item attributes and user preferences. It offers transparency by linking suggestions to content features and allows adaptability to diverse user tastes through item similarity \cite{lops2011content}.
Collaborative filtering generates recommendations by grouping users based on similar interaction patterns \cite{amoretti2017utravel}. Unlike content-based approaches focused on item attributes \cite{chen2010short}, it relies on user similarity \cite{nilashi2015clustering}, though it faces limitations like the cold-start problem when user data is insufficient.

\subsubsection{Machine Learning Methods}

While filtering methods offer adaptability, they often struggle to capture complex behavioural patterns due to reliance on predefined rules. Machine learning (ML) approaches address this limitation through data-driven pattern recognition, offering greater scalability and flexibility in both supervised and unsupervised settings. These models enable more nuanced and precise user profiling, especially when combined with natural language inputs.
Unsupervised ML techniques incorporating Natural Language Processing (NLP) have been applied to social network data for interest profiling \cite{bhargava2015unsupervised}. By leveraging both explicit and implicit signals—such as named entity recognition and sentiment analysis—these models capture user interests with greater granularity.
In supervised settings, neural networks have shown strong capabilities in feature learning and generalization. For example, deep-contextualized word embeddings have been used to detect bots on Twitter, using combinations of GloVe, ELMo, and bidirectional LSTMs to extract semantic and contextual cues from user-generated content \cite{heidari2020deep}. Similarly, emotional profiling through multi-layer co-occurrence networks has been applied to social media data, using word-hashtag relationships to build semantic representations \cite{stella2020lockdown}.
Despite these advances, ML models—particularly deep learning—still face challenges. Their black-box nature often limits interpretability, and their performance is sensitive to data quality and training bias. Moreover, the increased complexity of these models necessitates careful hyperparameter tuning and substantial computational resources for optimal performance.

\subsection{Persona development techniques}

Different from profiles as factual accounts presenting individuals, personas serve as an abstract representation of user groups with similar behaviours and preferences. We categorized the efforts in the persona development literature according to the sequence of their appearance in the following.

\subsubsection{Qualitative Personas}

Persona development was initially grounded in qualitative methods, often relying on small-scale data from interviews and questionnaires \cite{banerjee2004face}. These personas are manually crafted to capture in-depth, human-centric insights into user behaviours and needs.
A prominent qualitative technique is Manual Persona Development (MPD), where user segmentation is typically conducted using tools like affinity diagrams \cite{lindgren2007using}. While effective for capturing nuanced understanding, this manual process becomes increasingly subjective and prone to bias as the volume of data grows.
Although enhancements such as incorporating tacit knowledge from cross-functional teams \cite{mahamuni2018concise} and applying Latent Semantic Analysis for improved data interpretation \cite{miaskiewicz2008latent} have been proposed, such techniques still face limitations at scale. In contrast, data-driven approaches offer greater efficiency and objectivity for large-scale persona generation.

\subsubsection{Quantitative Personas}

Quantitative methods apply statistical analysis to large-scale datasets to uncover generalized user patterns and behaviours across broad populations \cite{goodman2018evaluating}. Building on this foundation, hybrid approaches that combine quantitative and qualitative data have gained traction \cite{tu2010combine}, offering both statistical rigor and contextual depth.
For instance, conjoint analysis has been used in software engineering to prioritize user-preferred features, resulting in better-aligned product designs \cite{aoyama2005persona}. In a different context, natural language processing techniques have been applied to film character dialogues to identify archetypes through textual pattern recognition \cite{bamman2013learning}. 
Together, these studies reflect a broader shift toward data-driven persona development, illustrating how quantitative techniques can enhance behavioural insight extraction and support more scalable persona creation.

\subsubsection{Automatically generated personas}
Automatically generated personas have been used to streamline the understanding of user behaviours by leveraging data analytics, offering a scalable alternative to the time-intensive and subjective nature of manual persona development.
Machine learning techniques such as K-means and Partitioning Around Medoids (PAM) clustering have been applied in Data-Driven Persona Development (DDPD) to enable scalable and adaptive modelling, particularly for large, real-time datasets \cite{hou2020method, salminen2021survey}.
Deep learning classifiers have also been combined with statistical techniques like Non-negative Matrix Factorization (NMF) to enhance Automatic Personality Detection (APD) \cite{salminen2020enriching}. In this approach, dimensionality reduction through NMF enables latent feature extraction, allowing deep learning models to focus analysis and improve persona relevance and accuracy.
Such automated approaches have also been applied to pattern recognition in domains like cybersecurity, for example, in theme analysis of online behaviour \cite{kim2019innocent}.
However, the limitations of ML-based persona development are consistent with those in the profiling domain. These include data-driven bias, a lack of interpretability due to their black-box nature, and a dependence on large training datasets.

\subsection{Feature Engineering}

\subsubsection{Textual Features} 

Various text representation methods have been used in NLP tasks to support effective profiling. Simpler models, such as Bag-of-Words (BoW) count word occurrences \cite{zainab2020detecting}, while more robust techniques, like TF-IDF, additionally weight words based on their relevance within documents \cite{zainab2020detecting, rezvani2020linking}.
N-gram encoding has been applied by mapping text to binary vectors based on a predefined dictionary, where each element reflects the presence or absence of specific sequences \cite{bandhakavi2017lexicon, eichstaedt2018facebook, majumder2017deep}. To reduce computational complexity, sentimentally neutral words may be excluded using expert judgment or sentiment analysis tools.
Social media and online platforms allow informal self-expression, often involving emoticons and slang. Since emotional understanding is crucial for persona modeling, replacing such informal elements with descriptive terms has been shown to improve contextual and emotional interpretation. The effectiveness of this transformation in enhancing model performance has been demonstrated in several studies \cite{afzoon2021enabling, kumar2019anxious}.

\subsubsection{Contextualized Features}

Contextualization embeds information within its broader environment, enhancing relevance and understanding by accounting for the diverse factors that shape user behaviour.
NMF has been applied to aggregated social media data to identify distinct user segments for persona generation \cite{jung2017persona}. By incorporating demographic attributes, this approach emphasizes the value of contextualized features in improving persona representation for user experience and product targeting.
Metadata integration from platforms such as YouTube Analytics and Google Analytics has been employed to support real-time, data-driven persona development \cite{salminen2018personas}. This enables dynamic personas that adapt over time to reflect evolving user behaviours and preferences.
In another study, a contextualized method for developing library user personas was implemented using usability tests, interviews, and direct observations \cite{sundt2017user}. An iterative process involving staff feedback was used to refine the personas and maintain their relevance to actual user experiences.
Although these approaches have advanced contextualization in persona development, integrating long-term profiling remains a promising direction. Furthermore, recent NLP models allow complex data analysis beyond traditional methods, enhancing both scalability and the extraction of deep behavioural insights.

\section{Model Overview}\label{sec3}

\begin{figure*}[!h]
\centering
  \includegraphics[width=\textwidth]{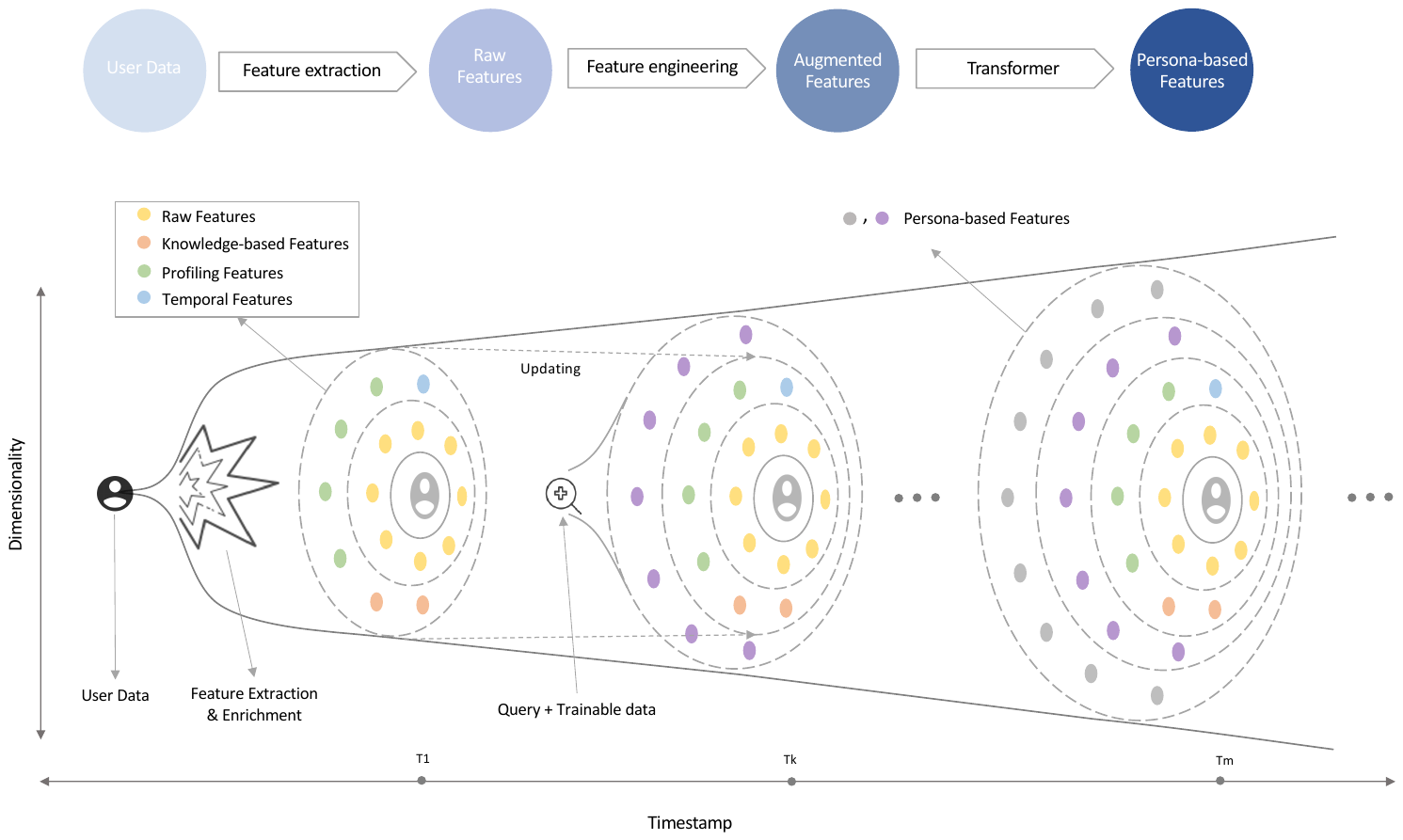}
  \caption{Overview of the ExBigBang framework illustrating the transformation of raw user data into enriched representations for persona classification.}
  \label{fig:model-overview-img}
\end{figure*}

This section introduces the ExBigBang framework, which integrates user profiling and persona classification through a hybrid text-tabular transformer. As illustrated in Fig.~\ref{fig:model-overview-img}, the pipeline transforms raw data into enriched representations via feature engineering, where textual inputs are contextualized using metadata, domain knowledge, and profile features. The following subsections describe the feature engineering process, profiling strategy, and transformer architecture in detail.

\subsection{Feature Engineering}

The creation of a text-tabular feature set is crucial for tuning the hybrid transformer, requiring both textual and contextual features to support accurate and explainable persona classification.

\subsubsection{Textual encoding}

To evaluate the impact of contextualization, baseline models use standard encoding methods. TF-IDF is applied in the ML model, while a word-level tokenizer is used in the deep neural network, serving as conventional textual encodings. In contrast, the hybrid transformer leverages a pre-trained Bidirectional Encoder Representations from Transformers (BERT) model, fine-tuned on the training data to capture deeper semantic relationships across user contexts. Parameter configurations for all encoding methods are provided in the experiment section.

\subsubsection{Contextual Features}

\begin{figure*}[h]
  \centering
    {\includegraphics[width=0.27\textwidth]{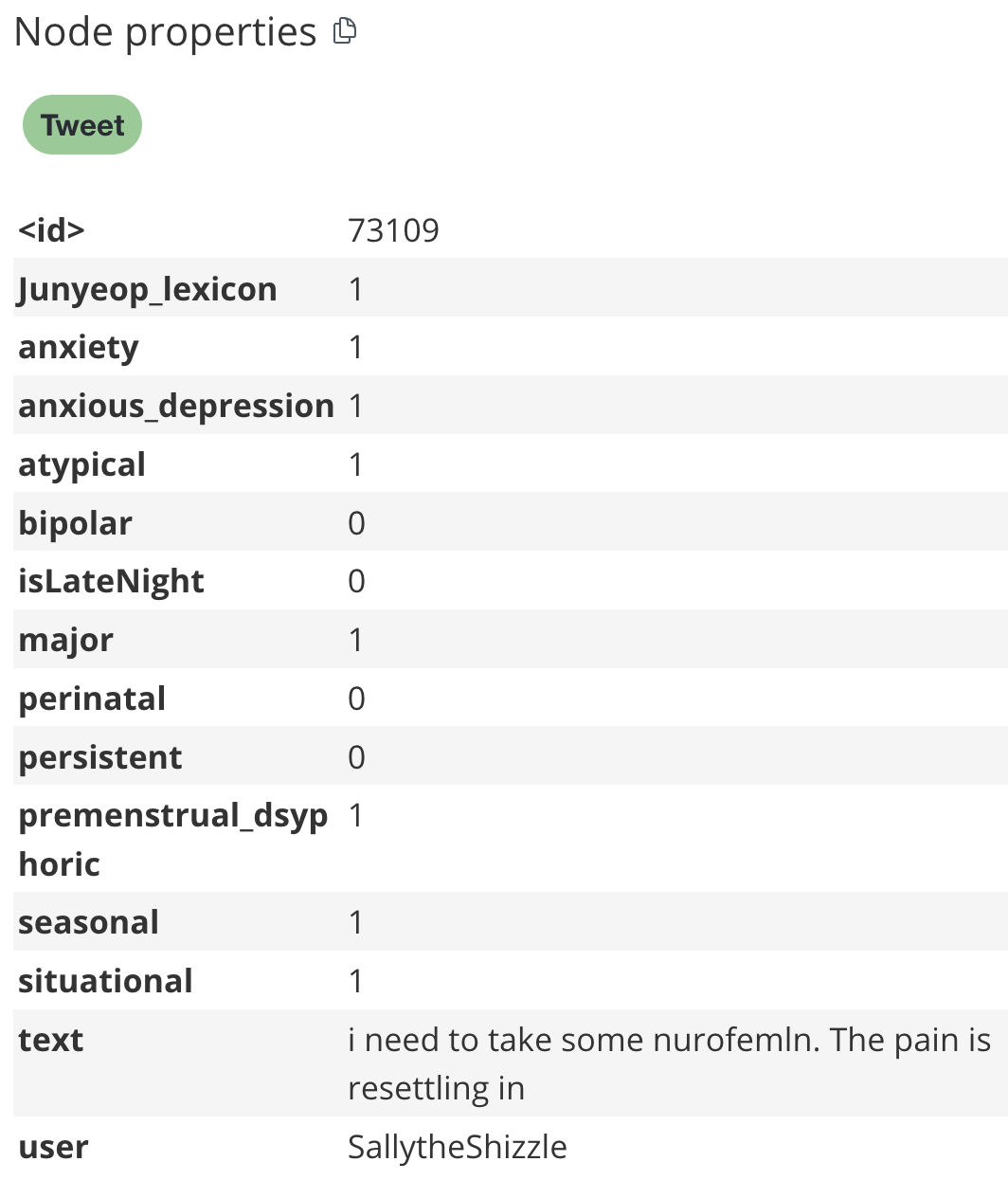}}
    \hfill
    {\includegraphics[width=0.36\textwidth]{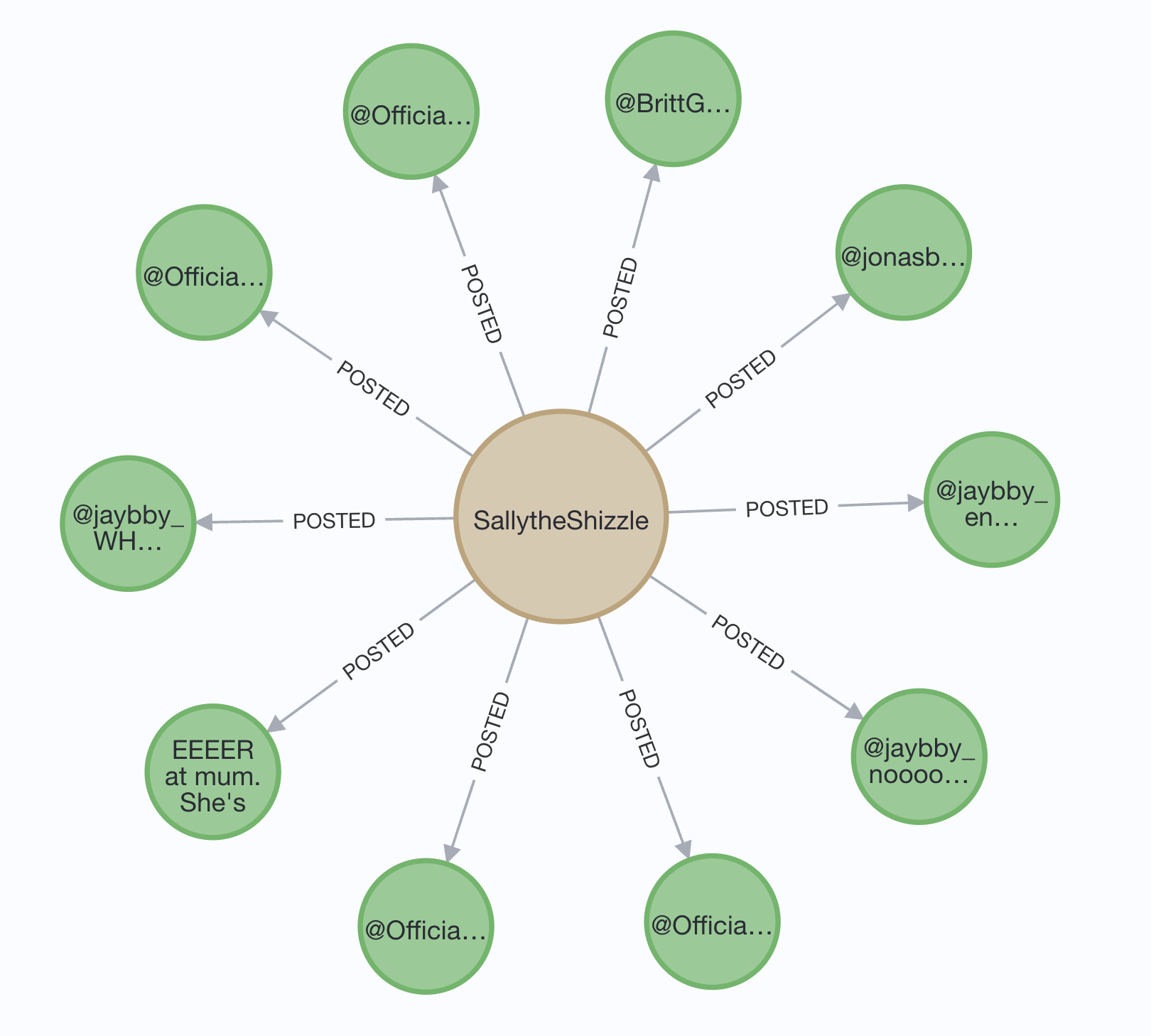}}
    \hfill
    {\includegraphics[width=0.27\textwidth]{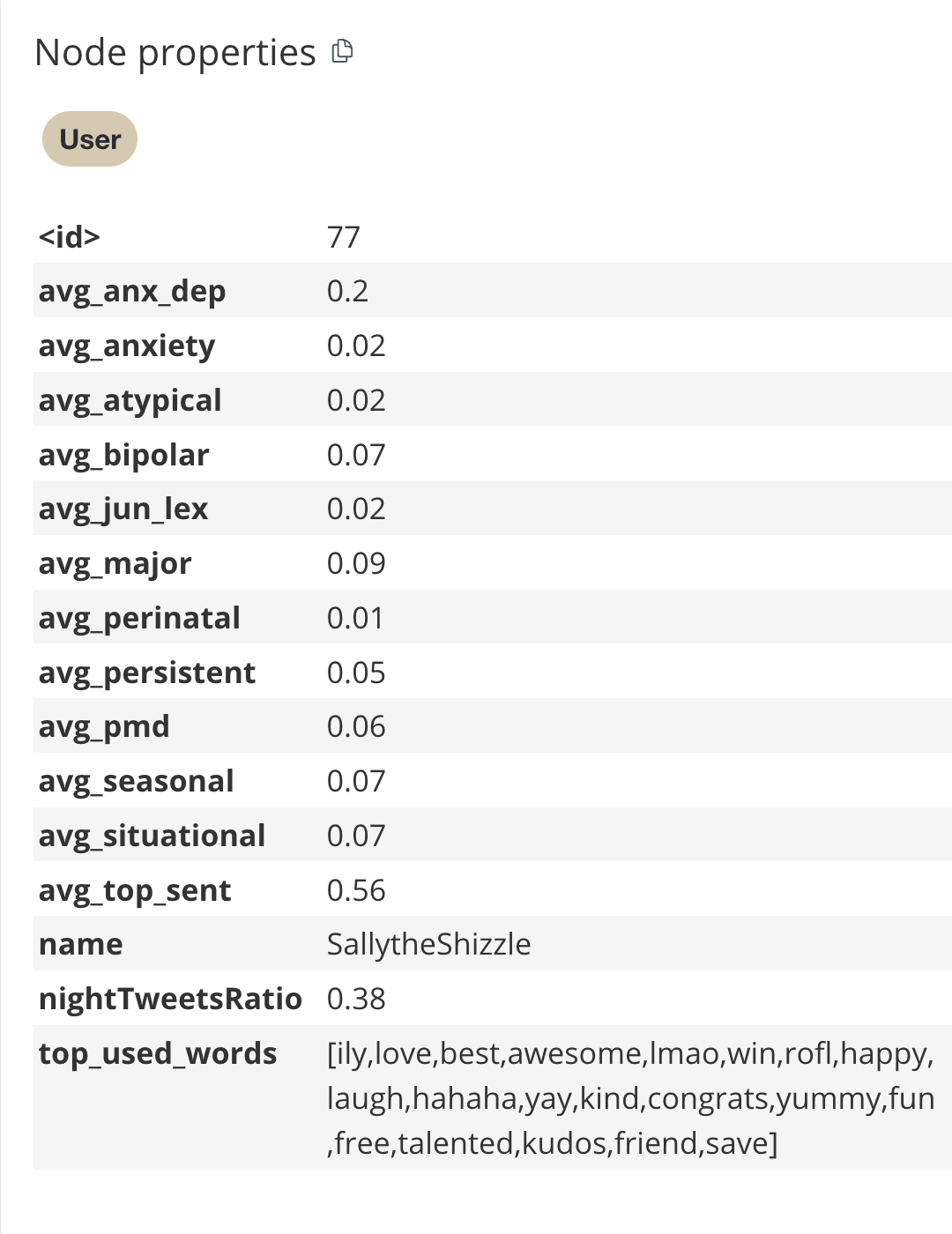}}
  \caption{A screenshot of the user-based contextualized profiling with individual profiled tweets on the left, an annotated user-centred graph in the middle, and the contextualized user profile on the right.}
  \label{fig:profiling}
\end{figure*}

Contextual features offer additional cues for task comprehension and help align model inference with human interpretation. Since this study focuses on behavioural analysis, contextualized features are constructed from the same domain. These include temporal, knowledge-based, and sentiment-based attributes.

A temporal feature, `isLateNight`, is used to enhance behavioural analysis related to insomnia. The 1–6 a.m. interval is identified as distinctive based on its correlation with target labels. This feature is also applicable to detecting other sleep-related issues.
Domain knowledge is incorporated using two behaviour-related lexicons: Junyeop\_lex \cite{cha2022lexicon} and a lexicon from \cite{afzoon2021enabling}, which extend tweet node attributes (Fig.~\ref{fig:profiling}).
Sentiment scores are calculated based on frequently used words in each user's text. This is done on aggregated profile data to prevent leakage. It is the only feature extracted directly from the user's profile and primarily supports model explainability.
To improve contextualization, user profiling captures long-term patterns and behavioural shifts, supporting transparency. As shown in Fig.~\ref{fig:profiling}, contextualized data are inserted as nodes, and users are connected through a shared name property, enabling time-aware updates and aggregation.
Profile-based features are generated through aggregated queries, including averaged lexicon scores, late-night activity ratios, and top-used words with compound sentiment values.
Profiling via sentiment and keyword aggregation automates feature enrichment, while domain knowledge integration supports customization across application contexts.

\subsection{Persona Development using Transformer}

\begin{figure*}
\centering
  \includegraphics[width=0.7\textwidth]{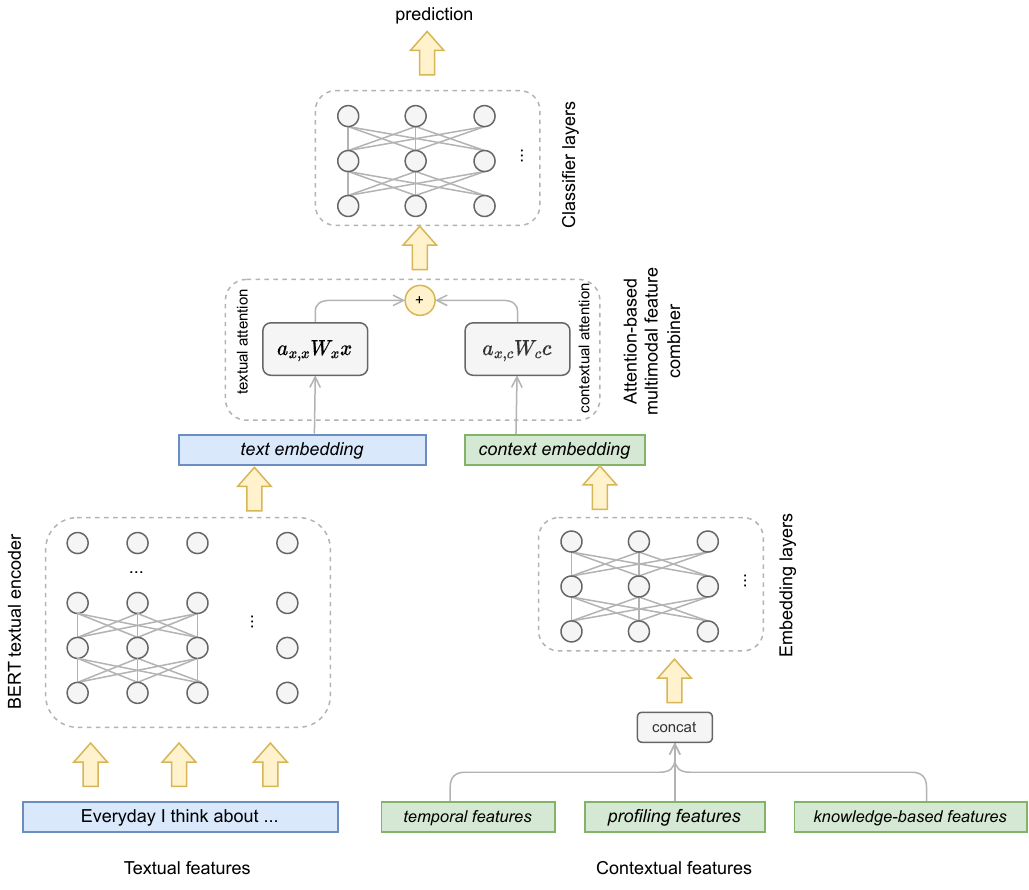}
  \caption{The hybrid Transformer model architecture used for persona development. It illustrates the attention deployment on textual and tabular contextual embeddings, followed by late fusion for final persona classification.}
  \label{fig:transformer-arch}
\end{figure*}

Transformer architecture has been adopted for its strengths in natural language understanding and sentiment analysis \cite{balci2021sentiment}, flexibility in NLP tasks \cite{chen2023separate}, and improved generalization via transfer learning \cite{amatriain2023transformer}.
As shown in Fig.~\ref{fig:transformer-arch}, a hybrid transformer model is used to derive persona-based features and enhance user profiling. The model receives classification queries related to profiling context, along with labeled training data (see Fig.~\ref{fig:model-overview-img}), and categorizes user profiles into personas based on behavioural similarity. These insights are then added back to the user profiles.
The transformer enables long-range dependency modeling, making it suitable for sentiment analysis of long-form user data. Attention mechanisms across different feature types—textual, categorical, and numerical—can be applied at various points in the architecture, either during embedding or post-encoding \cite{gu-budhkar-2021-package}.
In our implementation, attention-based summation is performed over BERT-encoded text, numerical, and categorical features before final classification:

\begin{equation}
m = a_{x,x} W_{x}x + a_{x,c} W_{c}c + a_{x,n} W_{n}n
\end{equation}
where
\begin{equation}
a_{i,j} =  \frac{exp(LeakyReLU( a^{T} [W_{i}x_{i} \mid\mid W_{j}x_{j}]))}
{\sum_{k\in\{x,c,n\}}exp(LeakyReLU( a^{T} [W_{i}x_{i} \mid\mid W_{k}x_{k}]))}
\end{equation}

Here, $m$ denotes the combined hybrid features, where $x$ is the text output from the transformer, $c$ the categorical, $n$ the numerical features, and $W$ a learnable weight matrix. The benefit of this hybrid attention mechanism is demonstrated in the results section.
As shown in Fig.~\ref{fig:transformer-arch}, textual data is first processed using BERT \cite{devlin2018bert}, while contextual inputs are embedded via a fully connected network. Their outputs are fused using the attention combiner, and passed to classification layers for final persona prediction.

\section{Experiments}\label{sec4}

This section evaluates the effectiveness of the proposed approach against baseline models in persona classification, using Accuracy, Precision, Recall, and F-score as evaluation metrics.
To analyze the contribution of contextual features beyond textual inputs, two baseline models were implemented: a conventional ML model and a deep learning model with constrained textual embeddings. This ensured a balanced comparison across contextual and textual features. Their results were then compared with the transformer-based model to assess the impact of model selection on classification quality.
The eXtreme Gradient Boosting (XGB) classifier was selected as the conventional ML baseline, given its common use in behavioural analysis contexts \cite{afzoon2021enabling, singh2019comparison, alsagri2020machine, deshpande2017depression, eichstaedt2018facebook}. Hyperparameters were optimized using Bayesian optimization \cite{joy2016hyperparameter}, with tuning applied to the learning rate, maximum depth, and number of estimators.
\begin{figure}[h]
  \centering
  \includegraphics[width=0.8\textwidth]{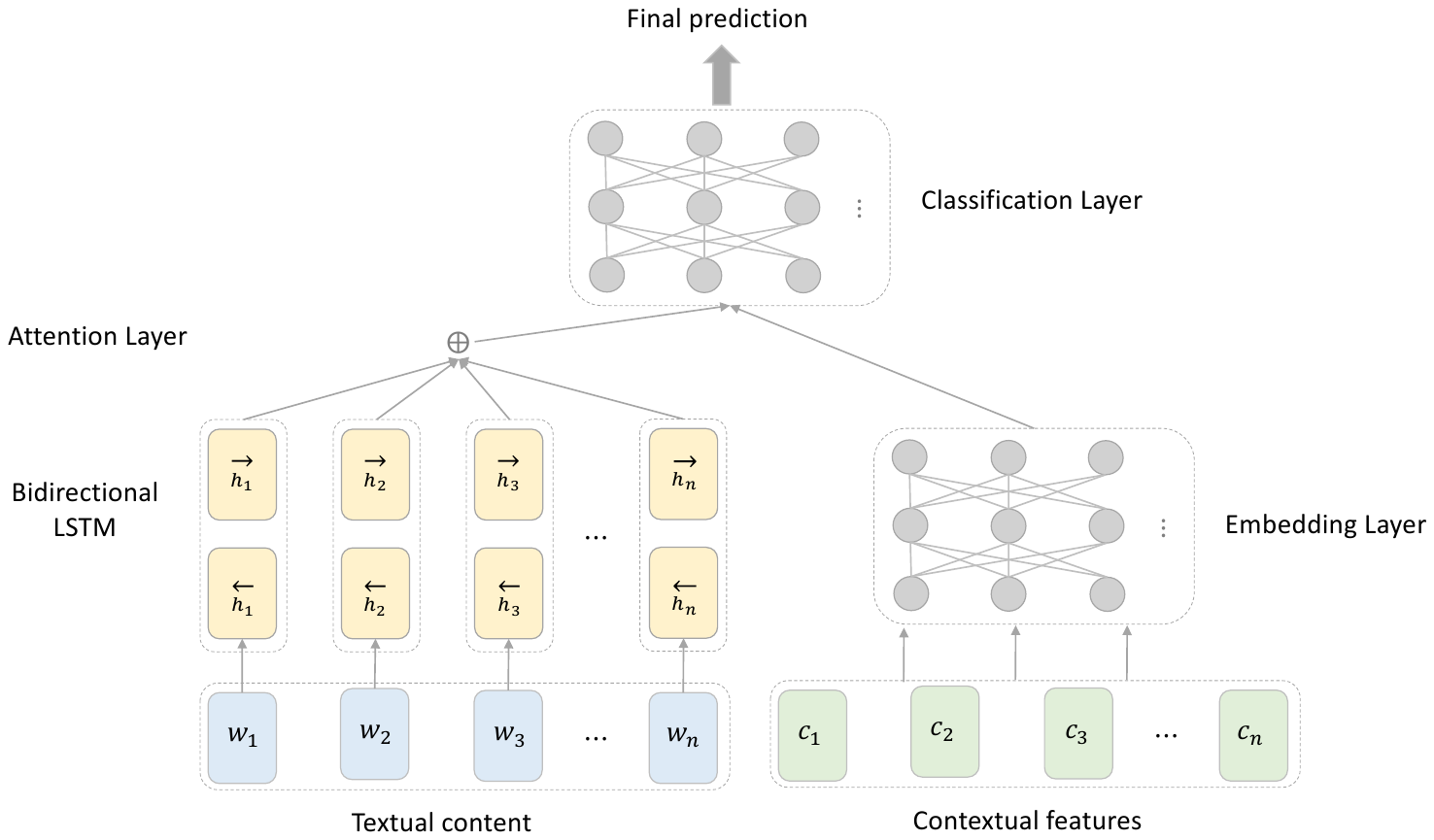}
  \caption{BiLSTM-Att NN architecture for processing textual content using a bidirectional LSTM and attention layer, combined with contextual feature embeddings.}
  \label{fig:deep-model-img}
\end{figure}
For the deep learning baseline, a hybrid DNN model was designed (Fig.~\ref{fig:deep-model-img}) inspired by prior behavioural analysis studies \cite{zhou2016text, rezvani2020linking, agrawal2018deep}. Textual content was preprocessed through tokenization and padding, followed by a BiLSTM layer for sequential pattern extraction. A single-head attention mechanism was applied to capture long-distance dependencies.
Contextual features were embedded via a dense layer and concatenated with the BiLSTM-attention output, forming the input to the final classification layer.

\subsection{Dataset}

\begin{figure}
\centering
  \includegraphics[width=0.5\textwidth]{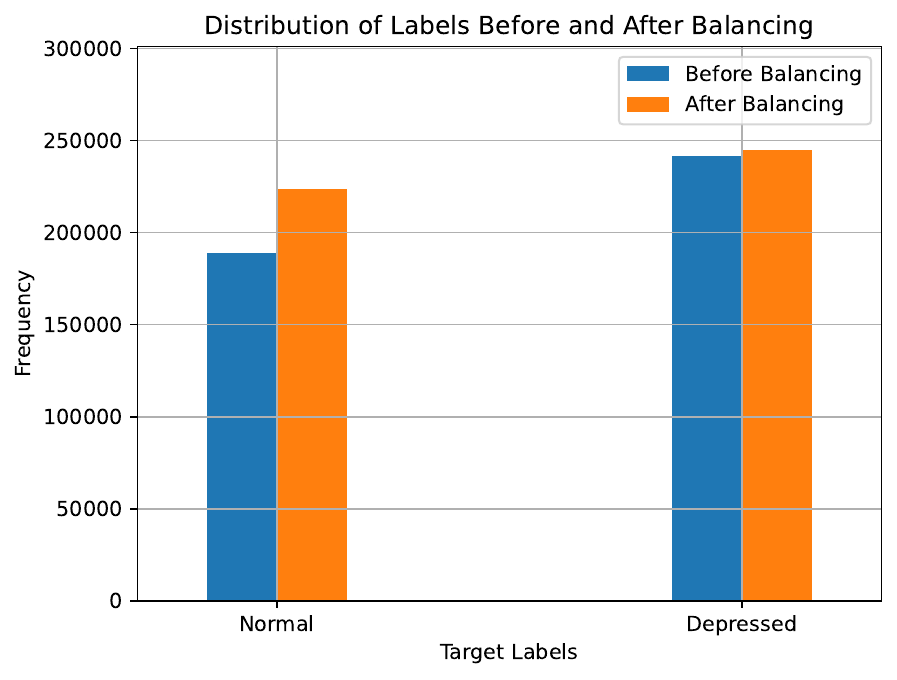}
  \caption{Data balancing using a user-based augmentation approach.}
  \label{fig:target_dist_analysis}
\end{figure}

Despite extensive review, no publicly available datasets were found that meet the dual requirements of sufficient textual volume for transformer-based training and comprehensive metadata for contextualization. This includes multimodal datasets compatible with the hybrid transformer's architecture. Additionally, baseline studies \cite{pattisapu2017medical, bharadwaj2018persona, kaul2020persona} do not reference any datasets that satisfy these criteria.
In alignment with baseline approaches, a social media dataset is selected for evaluation. The Twitter Sentiment140 dataset\footnote{\href{https://www.kaggle.com/kazanova/sentiment140}{Link to the Sentiment140 dataset}} is used for behaviour-oriented persona classification. It contains 1.6 million entries labeled as positive or negative. Although a neutral category is documented, no records in the dataset are explicitly marked as such.
For user-based record selection, tweets are grouped by user and sorted by tweet count. To fit within a 12 GB GPU memory constraint, users with fewer than 10 tweets are excluded, resulting in an initially unbalanced dataset of 430,000 tweets from 21,000 users. A 5\% class imbalance threshold is applied, selectively including additional users until balance is achieved, constrained by available memory. The final balanced dataset includes approximately 470,000 tweets from 27,000 users, as shown in Fig.~\ref{fig:target_dist_analysis}.
To parameterize the transformer, the classification query is set to infer depressive states. The dataset labels are used to fine-tune the transformer, while extracted features contribute to user profiling and feature construction.

\subsection{Baselines}

We benchmarked our method against both traditional and state-of-the-art methods. Our implementations of these baselines were used to cover both the proposed feature engineering methods and the employed ML solutions.

\subsubsection{Traditional Methods}
  
For feature extraction, we incorporated a variety of techniques recognized in the persona classification context. These included POS tagging \cite{pattisapu2017medical}, TF-IDF \cite{bharadwaj2018persona, afzoon2021enabling}, character n-grams, topic modeling with Linear Discriminant Analysis (LDA), and Word2Vec embeddings \cite{pattisapu2017medical} as our primary approaches. In terms of preprocessing, standard practices such as the removal of URLs, numbers, and punctuation were followed by lemmatization and stemming to refine the textual data \cite{bharadwaj2018persona, afzoon2021enabling}. The following are the machine learning solutions used by baselines, which are considered for an in-depth analysis of their effectiveness in persona development.

\begin{itemize}

  \item [$\bullet$] \textbf{Support Vector Machine (SVM)}\cite{bharadwaj2018persona, kaul2020persona, pattisapu2017medical}: Feature vectors are inputted into an SVM classifier with a linear kernel across three distinct setups: tabular-only, text-only (using each specified textual feature), and a combined text-tabular approach.
  \item [$\bullet$] \textbf{Naïve Bayes (NB)}\cite{bharadwaj2018persona, kaul2020persona}: In tabular-only, text-only, and text-tabular setups, features are fed to the Gaussian NB classifier.
  \item [$\bullet$] \textbf{Decision Tree (DT)}\cite{kaul2020persona}: In tabular-only, text-only, and text-tabular setups, features are fed to a DT classifier with Gini impurity.
  \item [$\bullet$] \textbf{Extreme Gradient Boosting (XGB)}\cite{afzoon2021enabling}: As an advanced decision tree variant proven superior in the related research, the XGB classifier with GBTree booster is applied in tabular-only, text-only, and text-tabular setups.
  
\end{itemize}

In the text-tabular setup, we only employ TF-IDF vectorization to structure the text-based features and lexicon-based and profiling features to craft our tabular features to enrich the analysis.

\subsubsection{Deep Learning (DL) based Methods}
Given the remarkable performance of RNN-based neural networks in a broad array of NLP tasks, we added NN architectures used in the persona development context to our baseline. We further expand our baseline by employing a Bidirectional LSTM alongside an attention layer, leveraging advanced NLP techniques to ensure a robust comparison with our ultimate transformer-based strategy.

\begin{itemize}

  \item [$\bullet$] \textbf{CNN-LSTM}\cite{pattisapu2017medical}: the classifier implemented by including a Convolutional layer followed by an LSTM layer and a binary cross-entropy objective function for behavioural persona development.
  \item [$\bullet$] \textbf{Bi-LSTM + ATT}: To enhance the baseline, we implement a Bidirectional LSTM with an attention layer for a balanced comparison against our hybrid transformer. For embeddings, we explore both Keras-based\footnote{\href{https://keras.io/}{https://keras.io/}} learned embeddings and pre-trained Glove\footnote{\href{https://nlp.stanford.edu/projects/glove/}{https://nlp.stanford.edu/projects/glove/}} embeddings.
  
\end{itemize}

We employed the Keras package to develop a BiLSTM model enhanced with attention mechanisms. This model standardizes tokenized text lengths with post-sequence padding and leverages a bidirectional LSTM followed by a custom single-head attention layer. ReLU and Sigmoid activations in dense layers facilitate the final classification.
In the Tabular-only setup, tabular features are input into an NN with an initial dense layer with ReLU activation, concluding with a sigmoid output layer for binary classification. In the text-tabular setup, the deep model employs textual embeddings instead of the textual vectorizing techniques used for the ML models.

\subsection{Implementation Setup}
In this section, we elaborate on the details of the setup for implementing benchmarks and the proposed hybrid transformer model, including their parameter settings.

\subsubsection{Domain Lexicon Design}

An extended lexicon adapted from a prior behavioural analysis study \cite{afzoon2021enabling} is used as domain knowledge. It includes various depression subtypes such as major, persistent, perinatal, seasonal, situational, and atypical, along with premenstrual dysphoric, anxiety, and anxious depression. For comparative analysis, the lexicon is augmented with a recently published resource, Junyeop\_lexicon, validated by a physiology specialist and developed in \cite{cha2022lexicon}.

\subsubsection{Transformer Model}

The transformer model is implemented using the Python `transformers` package\footnote{\href{https://pypi.org/project/transformers/}{https://pypi.org/project/transformers/}}, which supports state-of-the-art attention mechanisms widely used in NLP. A tokenizer from a pre-trained BERT model generates token-level embeddings, which are averaged to produce sentence-level representations. Contextual features are extracted and integrated with these embeddings. An attention mechanism merges the textual and contextual information, followed by a softmax-activated fully connected layer for classification.

Model training is performed on a system with a Tesla K80 GPU (12GB GDDR5 VRAM, 2496 CUDA cores), a 2.3 GHz hyper-threaded Xeon CPU, 12.6 GB RAM, and 33 GB of available storage.
All implementation details and resources are available on GitHub\footnote{\href{https://github.com/salehafzoon/BigBang}{https://github.com/salehafzoon/ExBigBang}}.

\subsubsection{Parameter Settings}
To determine suitable hyperparameters and define their search ranges, ChatGPT\footnote{ChatGPT by OpenAI was consulted for parameter selection.} was consulted. As a model trained on extensive domain-relevant literature, it supports selections that reflect current research trends and established best practices.
The hyperparameter values used throughout the analysis—including tuning for the Machine Learning, Deep Learning, and Transformer models—are detailed as follows:

\begin{enumerate}[label=(\arabic*)]
    \item Hyperparameter tuning setup using Bayesian optimization: max\_depth = (3, 10), gamma = (0,1),  learning\_rate = (0.01, 1), n\_estimators = (80, 150), subsample = (0.8, 1), colsample\_bytree = (0.8, 1).
    
    \item ML model setting: tokenizer = TF-IDF with a limit of 300 tokens, evaluation method = 3-fold cross-validation (due to hardware computational power limitations).
    
    \item Deep Learning model parameters: Keras tokenizer parameters: oov\_token = 'UNK', num\_words = 10000 (as its dictionary size). Sequence padding maximum length = 300 tokens, final layer activation function = "Sigmoid". model optimizer = "Adam" and loss function = "binary\_crossentropy".

    \item Transformer parameters: tokenizer = 'bert-base-uncased', activation function = 'relu', gating\_beta = 0.2, mlp\_dropout = 0.1, mlp\_division = 4, numerical\_transformer\_method = 'yeo\_johnson', categorical\_encode\_type = 'ohe', combine\_feat\_methods = ['text\_only', 'attention\_on\_cat\_and\_numerical\_feats'].
    
\end{enumerate}

\section{Results and Analysis}\label{sec5}
In this section, we present the results from our hybrid transformer model tailored for persona classification. We analyze the impact of the learning method and contextual features on output quality. Furthermore, we apply XAI to identify influential features in behavioural persona classification.
 
To ensure that the partially imbalanced dataset does not produce a remarkably biased prediction, our analysis in Table \ref{tab:bias-analysis} shows balanced accuracy (0.76) for both classes of tweets, despite the dataset's relative class imbalance.
\begin{table*}[h]
\centering
\caption{XGB Model Performance Metrics Highlighting Potential Bias in Negative vs. Positive Tweet Categories.}
\label{tab:bias-analysis}
\resizebox{0.35\textwidth}{!}{ 
\begin{tabular}{lcccl}
\hline
Class & precision & recall & f1-score  \\\hline
Negative     & 0.76      & 0.72   & 0.74       \\
Positive    & 0.76      & 0.80   & 0.78       \\
\\
\hline
\multicolumn{2}{|l|}{Accuracy} & \multicolumn{2}{l|}{0.76} \\
\multicolumn{2}{|l|}{Macro avg} & \multicolumn{2}{l|}{0.76} \\
\multicolumn{2}{|l|}{Weighted avg} & \multicolumn{2}{l|}{0.76} \\
\hline

\end{tabular}
}
\end{table*}

We compare the performance of our proposed transformer model against other baselines, presenting comprehensive results across various configurations using metrics such as accuracy, precision, recall, and F-score, as shown in Table \ref{tab:ds-res}.

\subsection{Impact of utilizing contextual features}

\begin{figure}[!t]
    \centering
    \begin{subfigure}[b]{0.48\textwidth}
        \centering
        \includegraphics[width=\textwidth]{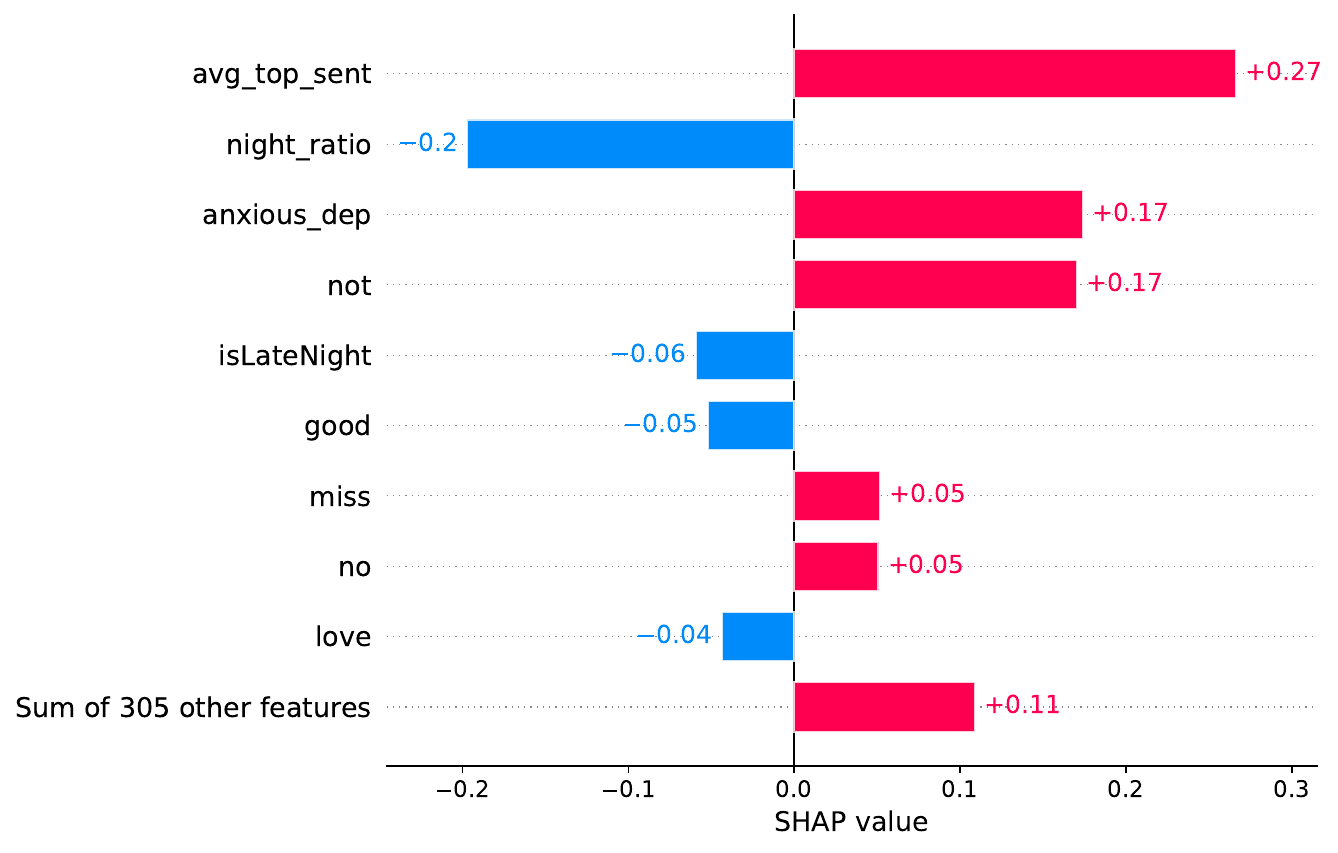}
        \caption{}
        \label{fig_first_case_shap}
    \end{subfigure}
    \hfil
    \begin{subfigure}[b]{0.48\textwidth}
        \centering
        \includegraphics[width=\textwidth]{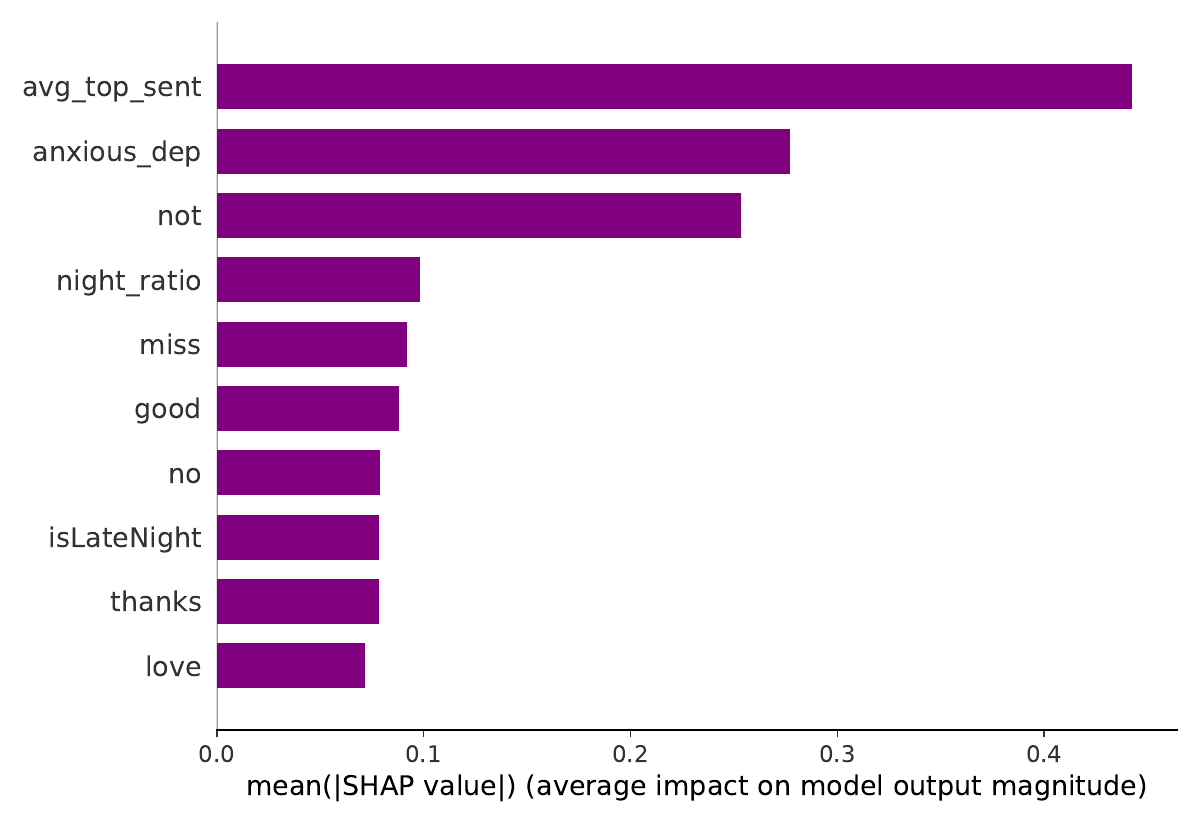}
        \caption{}
        \label{fig_second_case_shape}
    \end{subfigure}
    \caption{Analysis of feature importance: (a) Local SHAP value plot for a sample record and (b) Global plot showing average impact across all records.}
    \label{fig:shap_plots}
\end{figure}

As a hybrid feature engineering phase capable of fusing textual and contextual features can theoretically help comprehend the model's insight extraction, we analyse this factor in this section.
\begin{table*}[h]
\centering
\caption{Comparative results of baseline models and the proposed Transformer model across three different input setups. $\ast$ indicates that our transformer results are statistically significant ($p < 0.05$). Emb refers to Embedding.}
\resizebox{0.85\textwidth}{!}{
\begin{tabular}{
  l 
  l 
  S[table-format=2.2] @{${}\pm{}$} S[table-format=1.3] 
  S[table-format=2.2] @{${}\pm{}$} S[table-format=1.3] 
  S[table-format=2.2] @{${}\pm{}$} S[table-format=1.3] 
  S[table-format=2.2] @{${}\pm{}$} S[table-format=1.3] 
}
\toprule
& \multicolumn{1}{l}{Model} & \multicolumn{2}{c}{Accuracy} & \multicolumn{2}{c}{Precision} & \multicolumn{2}{c}{Recall} & \multicolumn{2}{c}{F1-Score} \\
\midrule
\multirow{2}{*}{Tabular Only} & NB  & 0.60 & 0.000 & 0.58 & 0.000 & 0.89 & 0.002 & 0.70 & 0.001 \\
                              & DT  & 0.65 & 0.000 & 0.66 & 0.000 & 0.70 & 0.006 & 0.68 & 0.003 \\
                              & SVM & 0.66 & 0.001 & 0.64 & 0.000 & 0.80 & 0.001 & 0.71 & 0.001 \\
                              & XGB & 0.67 & 0.001 & 0.66 & 0.002 & 0.79 & 0.004 & 0.71 & 0.001 \\
                              & Tabular Dense NN      & 0.66 & 0.002 & 0.76 & 0.018 & 0.64 & 0.001 & 0.67 & 0.002 \\
\midrule
\multirow{2}{*}{Textual Only} & NB + LDA\cite{kaul2020persona} & 0.55 & 0.002 & 0.56 & 0.012 & 0.62 & 0.246 & 0.56 & 0.012 \\
                              & NB + TF-IDF\cite{bharadwaj2018persona} & 0.72 & 0.001 & 0.72 & 0.001 & 0.76 & 0.000 & 0.74 & 0.001 \\
                              & SVM + POS\cite{pattisapu2017medical} & 0.57 & 0.001 & 0.58 & 0.002 & 0.69 & 0.009 & 0.63 & 0.003 \\
                              & SVM + LDA\cite{kaul2020persona} & 0.57 & 0.013 & 0.56 & 0.015 & 0.78 & 0.067 & 0.65 & 0.012 \\
                              & SVM + Char N-gram\cite{pattisapu2017medical} & 0.69 & 0.001 & 0.70 & 0.002 & 0.71 & 0.002 & 0.71 & 0.001 \\
                              & SVM + Word2Vec\cite{pattisapu2017medical} & 0.56 & 0.001 & 0.56 & 0.000 & 0.71 & 0.001 & 0.63 & 0.001 \\ 
                              & SVM + TF-IDF\cite{pattisapu2017medical, bharadwaj2018persona} & 0.74 & 0.000 & 0.74 & 0.000 & 0.78 & 0.000 & 0.76 & 0.000 \\
                              & DT + LDA\cite{kaul2020persona} & 0.54 & 0.019 & 0.56 & 0.013 & 0.52 & 0.070 & 0.64 & 0.045 \\ 
                              & XGB + TF-IDF\cite{afzoon2021enabling} & 0.74 & 0.000 & 0.73 & 0.000 & 0.79 & 0.001 & 0.76 & 0.000 \\
                              & CNN-LSTM\cite{pattisapu2017medical}  & 0.78 & 0.003 & 0.79 & 0.019 & 0.76 & 0.007 & 0.76 & 0.006 \\
                              & BiLSTM-Att + Glove  & 0.77 & 0.002 & 0.79 & 0.021 & 0.76 & 0.012 & 0.76 & 0.003 \\
                              & BiLSTM-Att + Learned Emb  & 0.79 & 0.001 & 0.80 & 0.007 & 0.78 & 0.003 & 0.78 & 0.002 \\
\midrule
\multirow{3}{*}{Text-Tabular} & DT    & 0.68 & 0.000 & 0.70 & 0.001 & 0.68 & 0.001 & 0.69 & 0.000 \\
                              & NB    & 0.70 & 0.002 & 0.71 & 0.002 & 0.73 & 0.000 & 0.72 & 0.001 \\
                              & SVM   & 0.73 & 0.001 & 0.72 & 0.001 & 0.80 & 0.001 & 0.75 & 0.001 \\
                              & XGB\cite{afzoon2021enabling}   & 0.76 & 0.001 & 0.76 & 0.000 & 0.79 & 0.002 & 0.78 & 0.001 \\
                              & CNN-LSTM \cite{pattisapu2017medical}   & 0.79 & 0.003 & 0.80 & 0.006 & 0.77 & 0.009 & 0.77 & 0.002 \\
                              & BiLSTM-Att + Glove  & 0.79 & 0.001 & 0.78 & 0.021 & 0.78 & 0.014 & 0.77 & 0.003 \\
                              & BiLSTM-Att + Learned Emb  & 0.80 & 0.001 & 0.80 & 0.008 & 0.79 & 0.007 & 0.79 & 0.001 \\
\midrule
& \textbf{Text-Tabular Transformer$\ast$} & \textbf{0.91} & \textbf{0.002} & \textbf{0.80} & \textbf{0.006} & \textbf{0.87} & \textbf{0.006} & \textbf{0.84} & \textbf{0.001} \\
\bottomrule
\end{tabular}
}
\label{tab:ds-res}
\end{table*}
In Table~\ref{tab:ds-res}, we compare three feature sets: tabular-only, textual-only, and text-tabular. The superiority of textual over tabular features highlights the dataset’s rich textual information and motivates the use of advanced NLP techniques. However, comparing text-tabular with text-only shows that integrating contextualized tabular features further improves performance. While advanced models like neural networks benefit from longer textual embeddings, simpler ML models show greater gains from the inclusion of contextual features. The relative impact of contextualization also depends on the balance between textual and tabular feature dimensions.

\begin{figure}
\centering
  \includegraphics[width=0.55\textwidth]{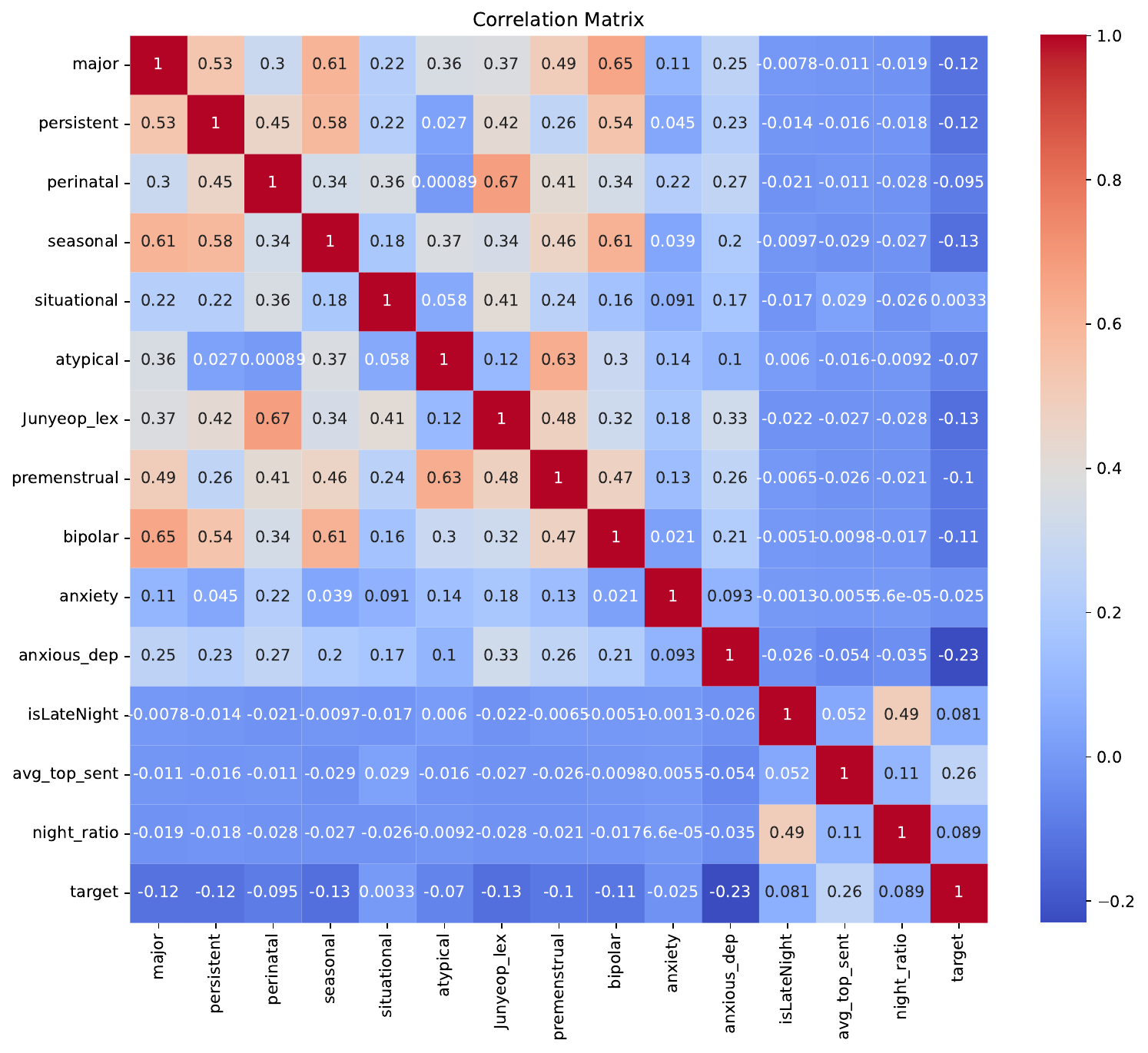}
  \caption{Feature set correlation matrix. night\_ratio is the ratio of late-night tweets per user and Junyeop\_lex is the lexicon from the study by Cha, J. et al. (2022)\cite{cha2022lexicon}.}
  \label{fig:corr_mat}
\end{figure}

To further examine this effect, SHapley Additive exPlanations (SHAP) \cite{lundberg2017unified} are used to interpret model predictions by quantifying each feature's contribution. SHAP computes the average model prediction across the dataset and assigns each feature a value indicating its effect. This analysis can be performed locally (per instance) or globally (across the dataset), as shown in Fig.~\ref{fig:shap_plots}. Local SHAP visualizations display each feature's positive or negative influence on a prediction using red and blue shading, while global SHAP summaries present average feature importance.
Fig.~\ref{fig:shap_plots} identifies \textit{avg\_top\_sent} as the most impactful contextual feature, reflecting alignment with the dataset’s domain. This sentiment-based profiling method is broadly applicable beyond mental health, including domains such as brand monitoring, customer feedback, and social media opinion mining \cite{dang2020sentiment, sharma2023analysis}. Calculating sentiment based on top-used keywords in a profile, rather than all individual posts, reduces feature-target correlation and prevents leakage, as shown in Fig.~\ref{fig:corr_mat}. The \textit{anxious\_dep} feature, derived from domain-specific knowledge, further demonstrates the framework's adaptability and effectiveness in content contextualization.

Temporal features designed for behavioural analysis, such as those addressing insomnia, also prove influential. Notably, the aggregated \textit{night\_ratio} feature outperforms the binary \textit{isLateNight}, illustrating how metadata becomes more informative when analyzed over time rather than as isolated points.
Finally, word-level analysis shows that terms like \textit{not}, \textit{no}, \textit{good}, \textit{thanks}, and \textit{love} often signal specific emotional states. For example, negative expressions like \textit{I don't feel good} are preprocessed to emphasize negation (e.g., \textit{not feel good}), enhancing sentiment detection. Conversely, terms such as \textit{good}, \textit{thanks}, and \textit{love} typically indicate positive sentiment, aiding the model in capturing emotional nuances in social media content.

A correlation heatmap (Fig.~\ref{fig:corr_mat}) is used to verify that no data or feature leakage occurs during feature extraction. The most influential feature, \textit{avg\_top\_sent}, shows a moderate correlation with the target label, while knowledge-based features (e.g., \textit{lexicon\_based}) show higher correlation but are retained to support deeper behavioural pattern analysis.
The heatmap also helps address possible misinterpretations in SHAP analysis, where interdependent features may cause over- or underestimation of importance. As shown, key contextual features such as \textit{avg\_top\_sent}, \textit{anxious\_dep}, and \textit{night\_ratio} exhibit low inter-feature correlation, reducing the likelihood of such distortion.

\subsection{Impact of Learning Model Selection}

Traditional learning methods rely solely on implicit, human-engineered features, making their performance highly dependent on domain expertise and the quality of those features.
In contrast, deep learning models automatically extract both explicit and implicit patterns, reducing the need for manual feature engineering. As shown in Table~\ref{tab:ds-res}, comparing classic ML, decision tree–based, and neural network models reveals how model complexity directly affects classification quality. However, deep learning’s sequential structure limits parallelization, motivating the exploration of more efficient model architectures.
%

\subsection{Impact of Transformer-Based Models on Textual and Contextual Features}

To unify earlier findings, a hybrid Transformer model is developed that applies attention mechanisms over contextual features and word-level embeddings. As shown in Table~\ref{tab:ds-res}, it significantly outperforms the baseline models.
In NLP, ML models depend on expert-crafted features due to their limited capacity to learn rich representations \cite{kumar2019anxious, bandhakavi2017lexicon, zainab2020detecting, eichstaedt2018facebook}. While MLPs can perform well, their training time is often prohibitive \cite{zhao2021artificial, muhlenbein1990limitations}. CNNs are effective for short texts by capturing word dependencies \cite{beheshti2020personality2vec, majumder2017deep, ying2019improving}, and LSTMs address vanishing gradients in RNNs but still struggle with complex language patterns \cite{rezvani2020linking}.
Transformers demonstrate superior performance in text understanding, attributed to their ability to model long-range dependencies \cite{lin2022survey}, making them particularly effective for persona classification in this study.
However, Transformers come with challenges. The self-attention mechanism greatly increases memory and computational demands, limiting dataset scalability on constrained hardware \cite{fan2020addressing}. Additionally, training requires tuning numerous hyperparameters and handling large datasets, increasing implementation complexity.

\subsection{Ablation Study}

\begin{figure}[h]
\centering
  \includegraphics[width=0.6\textwidth]{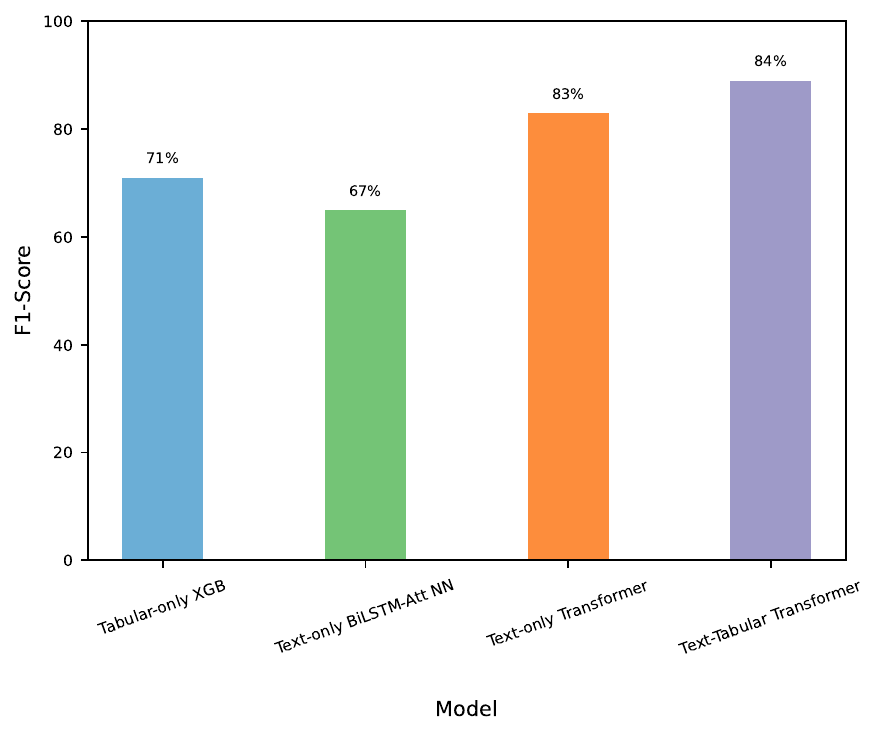}
  \caption{Comparative analysis of F1-Scores in an ablation study with the Text-Tabular Transformer as the baseline model.}
  \label{fig:ablation_plot}
\end{figure}

To analyze key contributors to model performance, an ablation study is conducted using the hybrid transformer as the baseline (Fig.~\ref{fig:ablation_plot}). Impactful contextual features are systematically removed to assess their influence. Despite being fewer in number than textual embeddings, these features still provide measurable gains. Their inclusion—even in a simple model like XGB—significantly enhances performance, emphasizing the value of domain-specific knowledge and the adaptability enabled by user profiling.
The modest yet consistent improvements of the text-tabular transformer over the text-only version further demonstrate the benefit of integrating high-quality contextual features with rich textual data.

\section{CONCLUSION AND FUTURE WORK}\label{sec6}

Personas serve as effective representations of user groups with shared preferences, behaviours, and goals. Persona development is essential for understanding target users, and when combined with user profiling, it enables deeper insight into behavioural dynamics.
This study introduces ExBigBang, a contextualized hybrid transformer framework that integrates user profiling and persona classification. Contextualization is achieved using domain knowledge, metadata, and profiling signals. Explainable AI is applied to interpret the influence of contextual features, offering insights into model performance and decision rationale.
Future work will explore multimodal data integration, richer metadata inclusion, and the use of knowledge graphs for deeper contextual understanding. Additional directions include transformer-driven storytelling and interactive dashboards for real-time analyst feedback.

\section*{Acknowledgments}
We acknowledge the Centre for Applied Artificial Intelligence at Macquarie University (Sydney, Australia) for funding this research.

\bibliographystyle{unsrt} 
\bibliography{main}

\end{document}